\documentclass[11pt]{article}
\usepackage{acl2016}
\usepackage{times}
\usepackage{url}
\usepackage{latexsym}

\aclfinalcopy % Uncomment this line for the final submission
\usepackage{graphicx}
\usepackage{color}
\usepackage{booktabs}
\usepackage{subcaption}
\usepackage{multirow}
\usepackage{amsmath}
\usepackage{amssymb}

\usepackage{pgfplots}
\pgfplotsset{compat=1.12}
\usepackage{tikz}
\usepackage{tkz-graph}
\usepackage{algorithmic}

\usepackage[ruled,vlined]{algorithm2e}
\newcommand\ngram{\ensuremath{n}-gram}
\newcommand\ngrams{\ensuremath{n}-grams}
\newcommand\witiii{WIT\raisebox{0.5ex}{\small3}}

\title{Findings of the 2016 WMT Shared Task\\ on Cross-lingual Pronoun Prediction}

\author{Liane Guillou \\ \normalsize{LMU Munich} \\ \normalsize{CIS} \\ \normalsize{liane@cis.uni-muenchen.de}
\And Christian Hardmeier \\ \normalsize{Uppsala University} \\ \normalsize{Dept. of Linguistics \& Philology} \\ \normalsize{christian.hardmeier@lingfil.uu.se}
\And Preslav Nakov \\ \normalsize{Qatar Computing Res.\ Inst.} \\ \normalsize{HBKU} \\ \normalsize{pnakov@qf.org.qa}
\AND Sara Stymne \\ \normalsize{Uppsala University} \\ \normalsize{Dept. of Linguistics \& Philology} \\ \normalsize{sara.stymne@lingfil.uu.se}
\And J\"{o}rg Tiedemann \\ \normalsize{University of Helsinki} \\ \normalsize{Dept. of Modern Languages} \\ \normalsize{jorg.tiedemann@helsinki.fi}
\And Yannick Versley \\ \normalsize{LinkedIn} \\ \normalsize{Dublin, Ireland} \\ \normalsize{yversley@gmail.com}
\AND Mauro Cettolo \\ \normalsize{Fondazione Bruno Kessler} \\ \normalsize{Trento, Italy} \\ \normalsize{cettolo@fbk.eu}
\And Bonnie Webber \\ \normalsize{ILCC, University of} \\ \normalsize{Edinburgh, Scotland, UK} \\ \normalsize{bonnie@inf.ed.ac.uk}
\And Andrei Popescu-Belis \\ \normalsize{Idiap Research Institute} \\ \normalsize{Martigny, Switzerland} \\ \normalsize{apbelis@idiap.ch}
}

\date{}

\begin{document}
\maketitle

\begin{abstract}
We describe the design, the evaluation setup, and the results of the 2016~WMT shared task on cross-lingual pronoun prediction. This is a classification task in which participants are asked to provide predictions on what pronoun class label should replace a placeholder value in the target-language text, provided in lemmatised and PoS-tagged form. 
%The task requires no specific MT expertise and is interesting as a machine learning task in its own right. 
% SS: I agree with APB and removed the above sentence
% APB: is this sentence really important for the abstract?
We provided four subtasks, for the English--French and English--German language pairs, in both directions. Eleven teams participated in the shared task; nine for the English--French subtask, five for French--English, nine for English--German, and six for German--English. 
Most of the submissions outperformed two strong language-model-based baseline systems, with systems using deep recurrent neural networks outperforming those using other architectures for most language pairs. % APB: added this about the results
\end{abstract}

\section{Introduction}
% TO BE COMPLETED BY: Andrei, Bonnie and Liane
% LKG: Motivate the need for the task
% LKG: Briefly explain the problem of cross-lingual pronoun prediction
% LKG: Reference the DiscoMT 2015 task
% LKG: Describe how the function of the pronoun in the source affects the translation options in the target
% LKG: Present examples of different pronoun functions and their translation?
Pronoun translation poses a problem for current state-of-the-art Statistical Machine Translation (SMT) systems \cite{LeNagardKoehn:2010,HardmeierFederico:2010,Novak:2011,Guillou:2012,Hardmeier:2014}. 

Problems arise for a number of reasons. 
In general, pronoun systems in natural language do not map well across languages, e.g., due to differences in gender, number, case, formality, or animacy/humanness, as well as due to differences in where pronouns may be used. 

To this is added the problem of \emph{functional ambiguity}, whereby pronouns with the same surface form may perform multiple functions \cite{Guillou:2016}. For example, the English pronoun ``it'' may function as an anaphoric, pleonastic, or event reference pronoun. An \emph{anaphoric} pronoun corefers with a noun phrase (NP). A \emph{pleonastic} pronoun does not refer to anything, but it is required by syntax to fill the subject position. An \emph{event reference} pronoun may refer to a verb phrase (VP), a clause, an entire sentence, or a longer passage of text. Examples of each of these pronoun functions are provided in Figure~\ref{fig:FunctionExamples}. It is clear that instances of the English pronoun ``it'' belonging to each of these functions would have different translation requirements in French and German.

\begin{figure}
\begin{center}
\begin{tabular}{|l@{\qquad}l|}
\hline
\textit{anaphoric}		& I have an \textbf{umbrella}. \textbf{It} is red.\\
\textit{pleonastic}		& I have an umbrella. \textbf{It} is raining.\\
\textit{event}			& He lost his job. \textbf{It} came as a total\\ 
& surprise.\\ \hline
\end{tabular}
\end{center}
\caption{\label{fig:FunctionExamples}Examples of three different functions fulfilled by the English pronoun ``it''.}
\end{figure}

% The translation of anaphoric pronouns is the most well-researched problem in pronoun translation for SMT. % APB: changed into the next line + added something
The problem of pronouns in machine translation has long been studied. In particular, for SMT systems, the recent previous studies cited above have focused on the translation of anaphoric pronouns. In this case, a well-known constraint of 
languages with grammatical gender is that agreement must hold between an anaphoric pronoun and the NP with which it corefers, called its \emph{antecedent}. The pronoun and its antecedent may occur in the same sentence (\emph{intra-sentential anaphora}) or in different sentences (\emph{inter-sentential anaphora}). Most SMT systems translate sentences in isolation, so inter-sentential anaphoric pronouns will be translated without knowledge of their antecedent and as such, pronoun-antecedent agreement cannot be guaranteed. The accurate translation of intra-sentential anaphoric pronouns may also cause problems as the pronoun and its antecedent may fall into different translation units (e.g., \ngram\ or syntactic tree fragment).

%APB: changed the beginning + added exception on "she" for ships
The above constraints start playing a role in pronoun translation in situations where several translation options are possible for a given source-language pronoun, a large number of options being likely to affect negatively the translation accuracy. In other words, pronoun types that exhibit significant \textit{translation divergencies} are more likely to be erroneously translated by an SMT system that is not aware of the above constraints.
For example, when translating the English pronoun ``she'' into French, there is one main option, ``elle'' (exceptions occur, though, e.g., in references to ships). 
However, several options exist for the translation of anaphoric ``it'': ``il'' (for an antecedent that is masculine in French) or ``elle'' (feminine), but also ``cela'', ``\c{c}a'' or sometimes ``ce'' (non-gendered demonstratives). 
% We might argue that pronouns with a greater number of translation options would be more difficult to translate than those with fewer. APB: idea moved above

% One way to model pronoun translation is to treat it as a cross-lingual pronoun prediction task. % APB: Changed the transition to the task description as follows:
The challenges of correct pronoun translation gradually raised the interest in a shared task, which would allow the comparison of various proposals and the quantification of their claims to improve pronoun translation. However, evaluating pronoun translation comes with its own challenges, as reference-based evaluation cannot take into account the legitimate variations of translated pronouns, or their placement in the sentence. Building upon the experience from a 2015 shared task, the WMT 2016 shared task on pronoun prediction has been designed to test capacities for correct pronoun translation in a framework that allows for objective evaluation, as we now explain.

\section{Task Description}
% TO BE COMPLETED BY: Andrei, Bonnie and Liane
% LKG: Describe the source-language pronouns and the target-language class labels for each subtask
% LKG: Justify restriction to subject position pronouns
% LKG: Justify the choice of target-language prediction labels (selected due to frequency of their alignment to the source-language pronouns)
% LKG: Mention why some prediction labels are present for one pair, and not for others. E.g. "you" for German--English
% LKG: The task has changed from DiscoMT 2015. Instead of raw text in the target language, we have PoS-tagged, lemmatised text (APB: yes, I think this is an essential point and needs explanation) 
% APB: Reformulated all the beginning of the section, to better motivate the design of the task and make the link with the DiscoMT 2015 task.

The WMT~2016 shared task on cross-lingual pronoun prediction is a classification task in which participants are asked to provide predictions on what pronoun class label should replace a placeholder value (represented by the token {\sc replace}) in the target-language text. It requires no specific Machine Translation (MT) expertise and is interesting as a machine learning task in its own right. Within the context of SMT, one could think of the task of cross-lingual pronoun prediction as a component of an SMT system. This component may take the form of a decoder feature or it may be used to provide ``corrected'' pronoun translations in a post-editing scenario.

The design of the WMT~2016 shared task has been influenced by the design and the results of a 2015 shared task \cite{Hardmeier:2015} organised at the EMNLP workshop on Discourse in MT (DiscoMT). The first intuition about evaluating pronoun translation is to require participants to submit MT systems --- possibly with specific strategies for pronoun translation --- and to estimate the correctness of the pronouns they output. This estimation, however, cannot be performed with full reliability only by comparing pronouns across candidate and reference translations because this would miss the legitimate variation of certain pronouns, as well as variations in gender or number of the antecedent itself. Human judges are thus required for reliable evaluation, following the protocol described at the DiscoMT~2015 shared task on \textit{pronoun-focused translation}. The high cost of this approach, which grows linearly with the number of submissions, prompted us to implement an alternative approach, also proposed in 2015 as \textit{pronoun prediction} \cite{Hardmeier:2015}. While
the structure of the WMT~2016 task is similar to the shared task of the same name at DiscoMT~2015, there are two main differences, one conceptual and one regarding the language pairs, as specified hereafter. % APB: I introduced this paragraph, I hope it looks OK to everyone

In the WMT~2016 task, participants are asked to predict a target-language pronoun given a source-language pronoun in the context of a sentence. In addition to the source-language sentence, we provide a lemmatised and part-of-speech (PoS) tagged target-language human-authored translation of the source sentence, and automatic word alignments between the source-sentence words and the target-language lemmata. 

In the translation, the words aligned to a subset of the source-language third-person subject pronouns are substituted by placeholders. The aim of the task is to predict, for each placeholder, the word that should replace it from a small, closed set of classes, using any type of information that can be extracted from the documents. In this way, the evaluation can be fully automatic, by comparing whether the class predicted by the system is identical to the reference one, assuming that the constraints of the lemmatised target text allow only one correct class (unlike the pronoun-focused translation task which makes no assumption about the target text). % APB: developed a bit the end

%SS: added an example and paragraph about it below. Remove if you don't like it!!
\begin{figure*}
\begin{center}
\begin{tabular}{|p{.9\textwidth}|}
\hline
ce OTHER~~~~ ce$|$PRON qui$|$PRON~~~~        It 's an idiotic debate . It has to stop .~~~~      REPLACE\_0 \^etre$|$VER un$|$DET d\'ebat$|$NOM idiot$|$ADJ REPLACE\_6 devoir$|$VER stopper$|$VER .$|$.~~~~      0-0 1-1 2-2 3-4 4-3 6-5 7-6 8-6 9-7 10-8
\\
\hline
\end{tabular}
\end{center}
\caption{\label{fig:examplePred}English--French example sentence from the development set with two {\sc replace} tags to be replaced by ``ce'' and ``qui'' ({\sc other} class), respectively. The French reference translation, not shown to participants, merges the two source sentences into one: ``C'est un d\'{e}bat idiot qui doit stopper.''}
\end{figure*}

Figure \ref{fig:examplePred} shows an English--French example sentence from the development set. It contains two pronouns to be predicted, indicated by {\sc replace} tags in the target sentence. The first ``it'' corresponds to ``ce'' while the second ``it'' corresponds to ``qui'' (equivalent to English ``which''), which belongs to the {\sc other} class, i.e., does not need to be predicted as is. This example illustrates some of the difficulties of the task: the two source sentences are merged into one target sentence, the second ``it'' becomes a relative pronoun instead of a subject one, and the second French verb has a rare intransitive usage.

The two main differences between the WMT~2016 and DiscoMT~2015 tasks are as follows. First, the WMT~2016 task introduces more language pairs with respect to the 2015 task. In addition to the English--French subtask (same pair as the DiscoMT~2015 task), we also provide subtasks for French--English, German--English and English--German. Second, the WMT~2016 task provides a lemmatised and PoS-tagged reference translation instead of the fully inflected text provided for the DiscoMT~2015 task. The use of this representation, whilst still artificial, could be considered to provide a more realistic SMT-like setting. SMT systems cannot be relied upon to generate correctly inflected surface form words, and so the lemmatised, PoS-tagged representation encourages greater reliance on other information from the source and target-language sentences.

The following sections describe the set of source-language pronouns and the target-language classes to be predicted, for each of the four subtasks. The subtasks are asymmetric in terms of the source-language pronouns and the prediction classes. 

The selection of the source-language pronouns and their target-language prediction classes for each subtask is based on the variation that is to be expected when translating a given source-language pronoun, i.e., the translation divergencies of each pronoun type. For example, when translating the English pronoun ``it'' into French, a decision must be made as to the gender of the French pronoun, with ``il'' and ``elle'' both providing valid options. Alternatively, a non-gendered pronoun such as ``cela'' may be used instead. The translation of the English pronouns ``he'' and ``she'' into French, however, does not require such a decision. These may simply be mapped one-to-one, as ``il'' and ``elle'' respectively, in the vast majority of cases. The translation of ``he'' and ``she'' from English into French is therefore not considered an \emph{interesting} problem and as such, these pronouns are excluded from the source-language set for the English--French subtask. In the opposite translation direction, the French pronoun ``il'' may be translated as ``it'' or ``he'', and ``elle'' as ``it'' or ``she''. As a decision must be taken as to the appropriate target-language translation of ``il'' and ``elle'', these are included in the set of source-language pronouns for French--English.

\subsection{English--French}
This subtask concentrates on the translation of subject-position ``it'' and ``they'' from English into French. The following prediction classes exist for this subtask (the class name, identical to the main lexical item, is highlighted in bold, but each class may include additional lexical items, indicated in plain font between quotes): % APB: added the parenthesis

\begin{itemize}\setlength\itemsep{-0.1em}
	\item \textbf{ce}: the French pronoun ``ce'' (sometimes with elided vowel as ``c'\thinspace'' when preceding a word starting by a vowel) as in the expression ``c'est'' (``it is'');
	\item \textbf{elle}: feminine singular subject pronoun;
	\item \textbf{elles}: feminine plural subject pronoun;
	\item \textbf{il}: masculine singular subject pronoun;
	\item \textbf{ils}: masculine plural subject pronoun;
	\item \textbf{cela}: demonstrative pronouns, including ``cela'', ``\c{c}a'', the misspelling ``ca'', and the rare elided form ``\c{c}'\thinspace'' when the verb following it starts with a vowel;
	\item \textbf{on}: indefinite pronoun;
	\item \textbf{OTHER}: some other word, or nothing at all, should be inserted.
\end{itemize}

\subsection{French--English}
This subtask concentrates on the translation of subject-position ``elle'', ``elles'', ``il'', and ``ils'' from French into English.\footnote{We explain below in Section~\ref{sec:SubjectFiltering} how non-subject pronouns are filtered out from the data.} % APB: added parenthesis 
The following prediction classes exist for this subtask:

\begin{itemize}\setlength\itemsep{-0.1em}
	\item \textbf{he}: masculine singular subject pronoun;
	\item \textbf{she}: feminine singular subject pronoun;
	\item \textbf{it}: non-gendered singular subject pronoun;
	\item \textbf{they}: non-gendered plural subject pronoun;
	\item \textbf{this}: demonstrative pronouns (singular), including both ``this'' and ``that'';
	\item \textbf{these}: demonstrative pronouns (plural), including both ``these'' and ``those'';
	\item \textbf{there}: existential ``there'';
	\item \textbf{OTHER}: some other word, or nothing at all, should be inserted.
\end{itemize}

\subsection{English--German}
This subtask concentrates on the translation of subject-position ``it'' and ``they'' from English into German. It uses the following prediction classes:

\begin{itemize}\setlength\itemsep{-0.1em}
	\item \textbf{er}: masculine singular subject pronoun;
	\item \textbf{sie}: feminine singular, and non-gendered plural subject pronouns;
	\item \textbf{es}: neuter singular subject pronoun;
	\item \textbf{man}: indefinite pronoun;
	\item \textbf{OTHER}: some other word, or nothing at all, should be inserted.
\end{itemize}

\subsection{German--English}
This subtask concentrates on the translation of subject position ``er'', ``sie'' and ``es'' from German into English. The following prediction classes exist for this subtask:

\begin{itemize}\setlength\itemsep{-0.1em}
	\item \textbf{he}: masculine singular subject pronoun;
	\item \textbf{she}: feminine singular subject pronoun;
	\item \textbf{it}: non-gendered singular subject pronoun;
	\item \textbf{they}: non-gendered plural subject pronoun;
	\item \textbf{you}: second person pronoun (with both generic or deictic uses);
	\item \textbf{this}: demonstrative pronouns (singular), including both ``this'' and ``that'';
	\item \textbf{these}: demonstrative pronouns (plural), including both ``these'' and ``those'';
	\item \textbf{there}: existential ``there'';
	\item \textbf{OTHER}: some other word, or nothing at all, should be inserted.
\end{itemize}

\section{Datasets}

\subsection{Data Sources}
% TO BE COMPLETED BY: Mauro and ???
% LKG: Describe each of the data sets used in training: TED, Europarl, News
% LKG: Provide stats in terms of sentences, tokens etc.
The training dataset comprises Europarl, News and TED talks data. The development and test datasets consist of 
TED talks. Below we describe the TED talks, the Europarl and News data, the method used for selecting the test datasets, and the steps taken to pre-process the training, development, and test datasets.

\subsubsection{TED Talks}

TED is a non-profit organisation that ``invites the world's most fascinating
thinkers and doers [...] to give the talk of their lives''. Its
website\footnote{http://www.ted.com/} makes the audio and the video of TED talks
available under the Creative Commons license. All talks are presented and
captioned in English, and translated by volunteers world-wide into many
languages.\footnote{As is common in other MT shared tasks, we do not give particular significance to the fact that all talks are originally given in English, which means that French--English translation is in reality a back-translation.}  % APB: added the parenthesis, ok?
In addition to the availability of (audio) recordings,
transcriptions and translations, TED talks pose interesting research challenges
from the perspective of both speech recognition and machine translation. 
Therefore,
both research communities are making increased use of them in building benchmarks.

TED talks address topics of general interest and are delivered to a live public audience
whose responses are also audible on the recordings.
% \footnote{The following overview of text characteristics is based on work by 	\newcite{Guillou:2014}.}  % APB: integrated this in the text below
The talks generally aim to be persuasive and to change the viewers' behaviour or beliefs. The genre of the TED talks is transcribed planned speech. 

As shown in analysis presented by Guillou et al.~\shortcite{Guillou:2014}, TED talks differ from other text types with respect to pronoun usage.
TED speakers frequently use first- and second-person
pronouns (singular and plural): first-person to refer to themselves and their
colleagues or to themselves and the audience, second-person to refer to the
audience, the larger set of viewers, or people in general. TED speakers often use the
pronoun ``they'' without a specific textual antecedent, in sentences such as
``This is what they think.''  They also use deictic and third-person pronouns
to refer to things in the spatio-temporal context shared by the speaker and the audience,
such as props and slides. In general, pronouns are common, and anaphoric
references are not always clearly defined.

For the WMT~2016 task, 
TED training and development sets come from the MT task of the 2015 IWSLT evaluation campaign~\cite{Cettolo:15:IWSLT}. The test set from DiscoMT~2015 \cite{Hardmeier:2015} was also released for development purposes.

\subsubsection{Europarl and News}

% LKG: Describe each of the data sets used in training: TED, Europarl, News

For training purposes, in addition to TED talks, the Europarl\footnote{http://www.statmt.org/europarl/} and News Commentary\footnote{http://opus.lingfil.uu.se/News-Commentary.php} corpora were made available. We used the alignments provided by OPUS, including the document boundaries from the original sources. For Europarl, we used version 7 of the data release and the News Commentary set refers to version 9. The data preparation is explained below.

% (\textcolor{purple}{WHO PREPARED SUCH DATA SHOULD PROVIDE SOME MORE INFO HERE})
% JT: is that enough about the corpora. We could have some more statistics but I don't think that this would be necessary here.

\subsection{Test Set Selection}

We selected the test datasets for the shared task from talks added recently to the TED repository that satisfy the following requirements:

\begin{enumerate}
\setlength{\parskip}{-1mm}
\item The talks have been transcribed (in English) and translated into both German and French.
\item They are not included in the training, development or test sets of the IWSLT
	evaluation campaigns, nor in the DiscoMT~2015 test set.
%    so they can be used as	held-out data with respect to those.
\item In total, they amount to a number of words suitable for evaluation purposes (some tens of thousands).
\end{enumerate}

Once we found the talks satisfying these criteria, we
automatically aligned them at the segment level. Then, we extracted a number of TED talks from the collection, following the criteria outlined in Section~\ref{sec:TestSetSelection}. Finally, we manually checked the sentence alignments of these selected TED talks in order to fix errors generated by either automatic or human processing.
Table~\ref{tab:tst2016stat} shows some statistics about the test datasets prepared for each subtask.

\begin{table}[ht]
\small
  \centering
  \begin{tabular}{crrr}
    \toprule
\multicolumn{1}{c}{subtask} &  \multicolumn{1}{c}{segs} & \multicolumn{2}{c}{\hspace*{0.2em}tokens\hspace*{0.2em}} \\
&    & \multicolumn{1}{c}{source} & \multicolumn{1}{c}{target}\\
    \midrule
English--French & 1,213 & 22,429 & 23,626 \\
French--English & 1,199 & 24,019 & 23,911 \\
English--German & 1,258 & 22,458 & 20,118 \\
German--English & 1,192 & 20,795 & 23,926 \\
    \bottomrule
  \end{tabular}
  \caption{Statistics about 2016 test datasets.}
  \label{tab:tst2016stat}
\end{table}

%\subsection{Test Set Selection Criteria} % APB: I remove this title because it duplicates the one included in the file "dataSets.tex" -- the two texts can flow one from another without sectioning
\label{sec:TestSetSelection}
% TO BE COMPLETED BY: Liane
% LKG: Describe the criteria used in selecting the TED talks for the test set of each subtask
% LKG: Mention that we have two test sets; one for translation from English, another for the translation into English
% LKG: Aimed to include more pronouns from the rare classes, and to have at least one pronoun for each class
% LKG: Selection was made prior to data preparation and some pronouns were filtered out. For one task, there are no pronouns of a particular task

\begin{table*}[ht]
\small
  \centering
  \begin{tabular}{llrrrrrrr}
    \toprule
	ID & Speaker & Segs & \multicolumn{2}{c}{Tokens} & & Segs & \multicolumn{2}{c}{Tokens}\\
    & & & English & French & & & English & German\\
    \midrule
    1541 & L. Kristine & 124 & 2,883 & 3,224 & & 124 & 2,883 & 2,614\\
    1665 & E. Schlangen & 48 & 1,027 & 1,087 & & 48 & 1,027 & 887 \\
    2155 & J. Howard & 174 & 3,943 & 3,794 & & 184 & 3,972 & 3,321 \\
    2175 & K. Gbla & 220 & 3,474 & 3,592 & & 249 & 3,475 & 3,110 \\
    2241 & P. Ronald & 161 & 2,870 & 3,104 & & 172 & 2,882 & 2,672 \\
    2277 & D. Hoffman & 225 & 3,736 & 3,837 & & 217 & 3,729 & 3,293 \\
    2289 & M. McKenna & 118 & 2,342 & 2,666 & & 121 & 2,338 & 2,207 \\
    2321 & Y. Morieux & 143 & 2,154 & 2,322 & & 143 & 2,152 & 2,014 \\
    \midrule
    & Total & 1,213 & 22,429 & 23,626 & & 1,258 & 22,458 & 20,118 \\
    \bottomrule
  \end{tabular}
  \caption{Test dataset documents: English to French/German.}
  \label{tab:EnglishToGermanFrench}
\end{table*}

\begin{table*}[ht]
\small
  \centering
  \begin{tabular}{llrrrrrrr}
    \toprule
	ID & Speaker & Segs & \multicolumn{2}{c}{Tokens} & & Segs & \multicolumn{2}{c}{Tokens}\\
    & & & French & English & & & German & English\\
    \midrule
    2039 & M. Gould Stewart & 105 & 2,567 & 2,443 & & 123 & 2,257 & 2,449 \\
    2140 & E. Balcetis & 127 & 2,725 & 2,541 & & 132 & 2,206 & 2,509 \\
    2151 & V. Myers & 151 & 2,803 & 2,918 & & 168 & 2,370 & 2,937 \\
    2182 & R. Semler & 235 & 4,297 & 4,530 & & 261 & 3,848 & 4,548 \\
    2194 & N. Burke Harris & 93 & 2,592 & 2,380 & & 105 & 1,977 & 2,369 \\
    2246 & A. Davis & 147 & 2,660 & 2,832 & & 103 & 2,347 & 2,805 \\
    2252 & E. Perel & 162 & 3,369 & 3,220 & & 163 & 3,162 & 3,226 \\
    2287 & C. Kidd & 179 & 3,006 & 3,047 & & 137 & 2,628 & 3,083\\
    \midrule
    & Total & 1,199 & 24,019 & 23,911 & & 1,192 & 20,795 & 23,926 \\
    \bottomrule
  \end{tabular}
  \caption{Test dataset documents: French/German to English.}
  \label{tab:GermanFrenchToEnglish}
\end{table*}

In total, we selected 16 TED talks for testing, which we split into two groups: 8 TED talks for the English to French/German direction, and 8 TED talks for the French/German to English direction. Another option would have been to create four separate groups of TED talks, one for each subtask. However, using a smaller set of documents reduced the manual effort in correcting the automatic sentence alignment of the documents.
%SS: is the above true? the sentence alignment is different for the two language pairs anyway, isn't it? 
% SS: I removed the below, since we did it separately for the two directions anyway, and since they were not exactly identical anyway, because of sentence segmentation.
% and the automatic filtering of subject position pronouns in English and German (see Section~\ref{sec:SubjectFiltering}).

The TED talks belonging to the test datasets are described in Tables~\ref{tab:EnglishToGermanFrench} and \ref{tab:GermanFrenchToEnglish}. The English texts used for the English--French and English--German subtasks are the same. Differences in alignment of the sentences leads to different segmentation of the parallel texts for the different language pairs. Minor corrections to the sentence alignment and to the text itself, which were applied manually, resulted in small differences in token counts for the same English TED talk when paired with the French vs.\ the German translation.

The TED talks in the test datasets were selected to include more pronouns from the rare classes. For example, for the English to French/German dataset, we wished to include documents that contained more feminine pronouns in the French and in the German translations. For the German/French to English dataset, we wished to include documents with more demonstrative pronouns in the English translations. The group of documents for the translation from English to French/German was balanced to ensure that the preference for rare pronouns was satisfied for both target languages.

\subsection{Data Preparation}
% TO BE COMPLETED BY: Sara and Jörg
% LKG: Describe the alignment of the training, dev and test data:

In order to extract pronoun examples, we first needed to align the data. We then extracted the pronoun examples based on the alignments. Finally, we filtered the examples in order to remove non-subjects. An innovation this year is the lemmatisation of the target data to remove the informative features coming from the inflections of the surrounding context. We used automatic lemmatisers and PoS taggers, and we further converted the PoS labels to 12 coarse universal PoS tags \cite{PETROV12.274}. For all languages in our dataset, we used TreeTagger \cite{Schmid:1994} with its built-in lemmatiser. The tagsets were then converted to universal PoS tags using publicly available mappings,\footnote{\url{https://github.com/slavpetrov/universal-pos-tags}} except for French, for which no appropriate mapping was available. In French, we clipped the morphosyntactic information from the base word class, which is separated by a colon (`:') in the tagset (e.g., \textit{VER:futu}, \textit{VER:impe} and all other verb tags would be reduced to \textit{VER}, thus only keeping the verb tag,
%and removing all other information, regarding for instance tense),
resulting in 15 tags. For German, we had to map pronominal adverbs to PROAV for the conversion to match the Tiger tagset used in the mapping to universal PoS tags.

% SS: Describe how the alignment+symmetrization was optimized: Sara
\subsubsection{Alignment Optimisation} % APB: let's put it one level below (subsub-) because it is under Data Preparation

Since we extract examples based on word alignments, we need good alignment precision in order not to extract erroneous examples, and good recall in order not to overgenerate the {\sc other} class. For the DiscoMT~2015 shared task, we explored this issue for English--French and found that GIZA++ model 4 and HMM with grow-diag-final-and symmetrisation gave the best results. For pronoun--pronoun links, we had an F-score of 0.96, with perfect recall and precision of 0.93 \cite{Hardmeier:2015}. This was slightly higher than for other links, which had an F-score of 0.92. 

For German--English, we explored this issue this year since it is a new language pair.
We used an aligned gold standard of 987 sentences from \cite{pado-lapata:05goldAlign}, which has been extensively evaluated by \newcite{stymne-tiedemann-nivre:2014:W14-33}. We used the same methodology as in 2015, and performed an evaluation on the subset of links between the pronouns we are interested in. We report precision and recall of links both for the pronoun subset and for all links, shown in Table~\ref{tab:deen_align}. 
The alignment quality is considerably worse than for French--English both for all links and for pronouns, but again the results for pronouns is better than for all links in both precision and recall. 

Across symmetrisation methods, HMM alignments give the best performance, especially for precision. The trade-off between precision and recall that holds for all links also applies to pronoun links. In the end, we decided to use HMM with intersection symmetrisation, since we believe that precision is more important than recall, in order not to add any false positive instances of the pronoun classes to our data. The lower recall will result in more examples from the {\sc other} class though. For English--French, we applied the same setup as last year using IBM Model 4 and the grow-diag-final-and symmetrisation heuristic. Similar to last year, we also perform backoff alignment with fast\_align in cases that are filtered out before running GIZA++ because of length and length-ratio restrictions of the parallel data.

\begin{table}
\small
  \centering
  \begin{tabular}{ccrrrr}
    \toprule
\multicolumn{1}{c}{Alignment} &  \multicolumn{1}{c}{Symmetrisation} & \multicolumn{2}{c}{All links} & \multicolumn{2}{c}{Pronouns}\\
&    & \multicolumn{1}{c}{P} & \multicolumn{1}{c}{R}   & \multicolumn{1}{c}{P} & \multicolumn{1}{c}{R} \\
    \midrule
Model 4 & \multirow{3}{*}{gdfa} & .75 & .79 & .82 & .88 \\
fast-align & & .69 & .73 & .80 & .81  \\ \cline{1-1}
\multirow{5}{*}{HMM} & &.80 & .73 & .87 & .85  \\ \cline{2-2}
 & gd & .81 & .70 & .89 & .78\\
 & gdf & .73 & .77 & .77 & .90\\
 & $\cup$ & .71 & .77 & .76 & .90 \\
 & $\cap$ & .92 & .61 & .92 & .74\\
    \bottomrule
  \end{tabular}
  \caption{Evaluation of German--English alignments for all links and pronouns using different alignment models and symmetrisation.}
  \label{tab:deen_align}
\end{table}

% SS: Describe the example selection: Jörg
\subsubsection{Example Selection} % APB: let's put it one level below (subsub-) because it is under Data Preparation

In order to select the acceptable target classes, we computed the frequencies of pronouns aligned to the ambiguous source-language pronouns based on the PoS-tagged training data. Using these statistics, we defined the sets of predicted labels for each language pair. Based on the counts, we also decided to merge small classes such as the demonstrative pronouns `these' and `those'.

Using these datasets, we identified examples based on the automatic word alignments. We include cases in which multiple words are aligned to the selected pronoun if one of them belongs to the set of accepted target pronouns. If this is not the case, we use the shortest word aligned to the pronoun as the placeholder token.

% APB: I edit a bit the first sentence below (and other words), because Christian also found this hard to understand: I hope I get it right!
% JT: perfect! thanks!
Unlike in 2015, we find a translation placeholder token for the unaligned pronouns using the following heuristic: we use alignment links of surrounding source-language words to determine the likely position for the placeholder token. 

We expand the window in both directions until we find a link. We insert the placeholder before or after the linked token, depending on whether the aligned source-language token is in left or right context of the selected pronoun. If no link is found in the entire sentence (an infrequent case), we use a position similar to the position of the selected pronoun within the source-language sentence. 
% This, however, does not happen very often.

% LKG: Describe the filtering out of non-subject position pronouns using the dependency parser - Sara
\subsubsection{Subject Filtering} % APB: let's put it one level below (subsub-) because it is under Data Preparation
\label{sec:SubjectFiltering}

The main interest of both the 2015 and the 2016 shared tasks has been on subject pronouns, and the pronoun sets have been selected with this in mind. However, several pronouns are ambiguous for the subject/object distinction. For the source datasets, this applies to English ``it'' and German ``es'' and ``sie''. In 2015, we ignored this issue, but this year we added a filtering step for the cases where English or German was the source language. We used automatic filtering for all datasets, and in addition, some manual filtering for the test dataset.

For the automatic filtering, we parsed the data using Mate Tools \nocite{Bohnet:2012} to perform joint PoS-tagging and dependency parsing. For the ambiguous pronouns, we then removed all pronoun instances that were not labelled as subjects, i.e., had the dependency label \emph{SBJ} for English or \emph{SB} for German. For French--English, no filtering was performed since all source pronouns are unambiguous subject pronouns.
Table \ref{tab:subject_filter_stats} shows how the subject filtering affected the IWSLT15 training set. For all languages, there was a large reduction for the {\sc other} class. For German--English, there were also large reductions for several other classes. Evaluations carried out after the shared task showed that this was mainly due to the dependency label \emph{EP}, which marks expletives, and which should not have been filtered away. This mainly affected translation from ``es gibt'' {/} ``there is'', and explains the large reduction of the \textit{there} class for German--English.

For the test dataset, we manually checked all of the pronouns that remained after the automatic filtering, in order to remove any remaining non-subjects. This showed that the performance of the parser for subjects was good and only a small amount of non-subjects remained, one for English--French, two for English--German, and six for German--English. We also noticed some issues with the casing of German ``Sie'', and changed it in four cases. Due to time constraints, we did not check the removed pronouns before releasing the data, but only for evaluation purposes afterwards. 

We checked all removed pronouns, 70 for English--German, 71 for English--French, and a sample of 70 pronouns for German--English, where many more pronouns were filtered away. For English as a source language, the filtering was very accurate, and there were only two instances for English--French and no instances for English--German where a subject pronoun had been removed erroneously. In both cases, the erroneous removal of the subject position pronoun was due to sentence segmentation issues. For German, though, 34 of the 70 removed pronouns were subjects. In 27 cases, they were labelled as expletives, as described above, which could easily be remedied. The remaining cases are indirect speech, relative clauses, or subordinate clauses, which appear to be more difficult for the parser than the English counterparts. Even so, the performance was acceptable also for German, with a much lower rate of non-subjects than before the filtering.

%SS: I don't know if we should keep this table in the paper, maybe it is too much information and could be removed? 
\begin{table*}
\small
  \centering
  \begin{tabular}{lrr lrr lrr}
    \toprule
\multicolumn{3}{c}{German--English} &  \multicolumn{3}{c}{English--German} & \multicolumn{3}{c}{English--French} \\
\multicolumn{1}{c}{word} & \multicolumn{1}{c}{before}   & \multicolumn{1}{c}{after} & \multicolumn{1}{c}{word} & \multicolumn{1}{c}{before}   & \multicolumn{1}{c}{after} & \multicolumn{1}{c}{word} & \multicolumn{1}{c}{before}   & \multicolumn{1}{c}{after} \\
    \midrule
he & 8,939 & 8,932 &       er &  2,217 & 2092   & ce  & 17,472 & 16,415 \\
she & 3,664 & 3,541 &      sie & 22,779 & 21041 & elle & 3,483 & 3,286 \\
it & 33,338 & 23,628 &     es &  26,923 & 21207 & elles & 3,305 & 3,276 \\
they & 18,581 & 17,896 &   man & 662 & 622     & il & 10,126 &  9,682\\
this & 1,479 & 983 &    OTHER & 32,197 & 21279 & ils  & 17,234 & 17,145 \\
these & 250 & 172 &          & &             & cela & 8,071 & 6,908 \\
there & 6,935 & 2,905 &        & &             & on & 1,713 & 1,549 \\
OTHER  & 30,751 & 18,102 &     & &             & OTHER & 27,530 & 11,226 \\
    \bottomrule
  \end{tabular}
  \caption{Number of pronouns for the different classes in the IWSLT15 data before and after filtering.}
  \label{tab:subject_filter_stats}
\end{table*}

\section{Baseline Systems}
\label{sec:BaselineSystems}
% TO BE COMPLETED BY: Yannick / Liane
% LKG: I've adapted the text from the DiscoMT 2015 paper and my understanding of the changes. The text in this section should be sanity-checked by other members of the team
The baseline system for each language pair is based on an \ngram\ language model. The architecture is similar to that used for the DiscoMT~2015 cross-lingual pronoun prediction task, but the systems are trained on lemmatised, PoS-tagged
data instead of raw, unprocessed text. Given that none of the systems submitted to the cross-lingual pronoun prediction task at DiscoMT~2015 were able to beat the baseline system, we deemed it suitable for re-use this year. 

We provided baseline systems for each subtask. Each baseline is based on a 5-gram language model trained on word lemmata,
%SS: no tags were used in the LM, were they??
constructed from news texts, parliament debates, and the TED talks of the training/development portions of the datasets. The additional monolingual news data comprises the shuffled news texts from WMT including the 2014 editions for German and English and the 2007--2013 editions for French. The German corpus contains a total of 46 million sentences with 814 million lemmatised tokens, English contains 28 million sentences and 632 million tokens, and French includes 30 million sentences with 741 million tokens.
%SS: I think we should write in more detail what data we used, and how big it is! JT?
% JT: Is this good enough?
% SS: yes!

The justification for using a baseline system based on a language model remains unchanged from the DiscoMT~2015 shared task. That is, the aim is to reproduce the most realistic scenario for a phrase-based SMT system.

The main assumption here is that the amount of information that can be extracted from the translation table is not sufficient or is inconclusive. As a result, the pronoun prediction would be influenced primarily by the language model. 

The baseline system fills the {\sc replace} token gaps by using a fixed set of pronouns (those to be predicted) and a fixed set of non-pronouns (which includes the most frequent items aligned with a pronoun in the provided test set) as well as the {\sc none} option (i.e., do not insert anything in the hypothesis). The baseline system may be optimised using a configurable {\sc none} penalty that accounts for the fact that \ngram\ language models tend to assign higher probability to shorter strings than to longer ones.

Two official baseline scores are provided for each subtask. The first was computed with the {\sc none} penalty set to an unoptimised default value of zero. The second was computed with the {\sc none} penalty set to an optimised value, which is different for each subtask. The {\sc none} penalty was optimised on the development set by a grid search procedure where we tried values between 0 and $-4$, with a step of 0.5. 

%SS: I could give the socres for different non-penalties for the dev and test sets? Would that be interesitng enough to add?? 

\section{Submitted Systems}
% TO BE COMPLETED BY: Liane and ???
% LKG: Briefly summarise the approach taken by each group
% LKG: Compare and contrast systems: try to look for similarities, differences
Eleven teams participated in the shared task, but not all teams submitted systems for all subtasks. Some teams also submitted second, contrastive systems for some subtasks. Ten of the groups submitted system description papers, which are cited hereafter. For the eleventh submission, {\sc UU-Cap}, no system description paper was submitted. Brief summaries of each submission, including {\sc UU-Cap}, are presented in the following sections.

\subsection{CUNI}
Charles University participated in the English--German and German--English subtasks \cite{CUNI:2016}. Each {\sc cuni} system is a linear classifier trained using a logistic loss optimised using stochastic gradient descent, implemented in the Vowpal Wabbit toolkit.\footnote{\url{https://github.com/JohnLangford/vowpal_wabbit/wiki}} In the primary submission, the training examples are weighed with respect to the distribution of the target pronouns in the training data, which aims at improving the prediction accuracy of less frequent pronouns. The contrastive submission does not weigh examples.

Before extracting the examples as feature vectors, the data is linguistically preprocessed using the Treex framework \cite{treex}. The source-language texts undergo a thorough analysis and are enriched with PoS tags, dependency syntax, as well as semantic roles and coreference for English. On the other hand, only grammatical genders are assigned to nouns in the target language texts. The system uses three types of features: the features based on the target-language model estimates provided by the baseline system, linguistic features concerning the source word aligned to the target pronoun, and approximations of the coreference and dependency relations in the target language.

Following the submission of the {\sc cuni} systems for English--German, an error was discovered in the merging of the classifier output into the test data file for submission. Fixing it yielded an improvement, with the contrastive system achieving recall of 51.74, and 54.37 for the primary system.

Except for the English wordlist with gender distributions by \newcite{bergsma-gennum}, only the shared task data was used in the {\sc cuni} systems.

\subsection{IDIAP}
The {\sc idiap} systems \cite{IDIAP:2016} focus on English--French using two types of target-side information: a target-side pronoun language model (PLM) and several heuristic grammar rules. The goal is to test how much a target-side only PLM can improve the translation of pronouns, without any knowledge of the source texts, i.e., by looking at target-side fluency only.

The rules are specifically constructed for predicting two cases: the French pronoun ``on'' and the untranslated pronouns. They detect the source and target patterns signalling the possible presence of such pronouns, which are not always correctly captured by SMT systems.

For predicting all of the other pronouns, the {\sc idiap} system relied solely on the scores coming from the proposed PLM model. This target-side PLM model uses a large target-language training dataset to learn a probabilistic relation between each target pronoun and the distribution of the gender-number of its preceding nouns and pronouns. For prediction, given each source pronoun ``it'' or ``they'', the system uses the PLM to score all possible candidates and to select the one with the highest score.

In addition to the PoS-tagged lemmatised data that was provided for the shared task, the \witiii\ parallel corpus \cite{Cettolo:2012}, provided as part of the training data at the DiscoMT~2015 workshop, was used to train the PLM model. Furthermore, a French PoS-tagger, Morfette \cite{Chrupala:2008}, was employed for gender-number extraction.

\subsection{LIMSI}
The {\sc limsi} systems \cite{LIMSI:2016} for the English--French task are linguistically-driven statistical classification systems. The systems use random forests, with few, high-level features, relying on explicit coreference resolution and external linguistic resources and syntactic dependencies. The systems include several types of contextual features, including a single feature using context templates to target particularly discriminative contexts for the prediction of certain pronoun classes, in particular the {\sc other} class.

The difference between the primary and contrastive systems is small. In the primary system, the feature value `number' is assigned by taking the number of the last referent in the English-side coreference chain. In the contrastive system, the value of `number' was taken directly from the English pronoun that was aligned with the placeholder: plural for ``they'' and singular for ``it''.

A number of tools and resources are used in the {\sc limsi} system. Stanford CoreNLP is used for PoS tagging, syntactic dependencies, and coreference resolution over the English text. The Mate Parser \cite{Bohnet:2012}, retrained on SPMRL 2014 data \cite{Seddah:2014} (dependency trees), and the Lefff \cite{Sagot:2010}, a morphological and syntactic lexicon (used for information on noun gender and impersonal adjectives and verbs), are both used for French.

\subsection{TurkuNLP}
\label{sec:SystemDescription:TurkuNLP}
The architecture for the {\sc TurkuNLP} system \cite{TurkuNLP:2016} is based on token-level sequence classification around the target pronoun using stacked recurrent neural networks.

The system learns token-level embeddings for the source-language lemmata, target-language tokens, PoS tags, combination of words and PoS tags and separate embeddings for the source-language pronouns that are aligned with the target pronoun. The network is fed sequences of these embeddings within a certain window to the left and to the right of the target pronoun. The window size used by the system is 50 tokens or until the end of the sentence boundary.

All of these inputs are read by two layered gated recurrent unit neural networks, except for the embedding for the aligned pronoun. All outputs of the recurrent layers are concatenated to a single vector along with the embedding of the aligned pronoun. This vector is then used to make the pronoun prediction by a dense neural network layer.

The primary systems are trained to optimise macro-averaged recall and the contrastive systems are optimised without preference towards rare classes.
The system is trained only on the shared task data and all parts of the data, in-domain and out-of-domain, are used for training the system.

\subsection{UEDIN}
The {\sc uedin} systems \cite{UEDIN:2016} for English--French and English--German are Maximum Entropy (MaxEnt) classifiers with the following set of features: tokens and their PoS tags are extracted from a context window around source- and target-side pronouns. $N$-gram combinations of these features are included by concatenating adjacent tokens or PoS tags. Furthermore, the pleonastic use of a pronoun is detected with NADA \cite{Bergsma:2011} on the source side. 

A Language Model (LM) is used to predict the most likely target-side pronoun, and then it is included as a feature. Another feature extracts the closest target-side noun antecedent (and its gender for German) via source coreference chains and word alignments. Additionally, the systems learn to predict NULL-translations (i.e., pronouns that do not have an equivalent translation). Experiments with linear-chain Conditional Random Fields (CRFs) treating pronouns of the same coreference chain as a sequence are conducted as well. All models are trained on a subset of the provided training data that has well-defined document boundaries in order to allow for meaningful extraction of coreference chains.

The MaxEnt classifiers consistently outperform the CRF models. Feature ablation shows that the antecedent feature is useful for English--German, and predicting NULL-translations is useful for English--French. It also reveals that the LM feature hurts performance.

\subsection{UHELSINKI}
The {\sc uhelsinki} system \cite{UHELSINKI:2016} implements a simple linear classifier based on LibSVM with its L2-loss SVC dual solver. The system applies local source-language and target-language context using the given tokens and PoS labels as features. Coreference resolution is not used, but additional selected items in the prior context are extracted to enrich the model. In particular, a small number of the nearest determiners, nouns and proper nouns are taken as possible antecedent candidates. The contribution of these features is limited even with the lemmatised target-language context that makes it harder to disambiguate pronoun translation decisions. The model performs reasonably well especially for the prediction of pronoun translations into English.

\subsection{UKYOTO}
The {\sc ukyoto} system \cite{UKYOTO:2016} is a simple Recurrent Neural Network system with an attention mechanism which encodes both the source sentence and the context of the pronoun to be predicted and then predicts the pronoun. The interesting thing about the approach is that it uses a simple language-independent Neural Network (NN) mechanism that performs well in almost all cases. Another interesting aspect is that good performance is achieved, even though only the IWSLT data is used. 

This indicates that the NN mechanism is quite effective. The only side effect is that the neural network overfits on the training and on the development datasets. In the future, the authors plan to use coreference resolution and system combination, which should help improve the performance.

\subsection{UUPPSALA}
%Since no suitable tools exist for the detection of event reference pronouns in English, 
The main contribution of the {\sc uuppsala-primary} system \cite{UUPPSALA:2016} for English--French is a Maximum Entropy classifier used to determine whether an instance of the English pronoun ``it'' functions as an anaphoric, pleonastic, or event reference pronoun. The classifier is trained on a combination of \emph{semantic}, based on lexical resources such as VerbNet \cite{Schuler:2005} and WordNet \cite{Miller:1995}, and frequencies computed over the annotated Gigaword corpus \cite{Napoles:2012}, \emph{syntactic}, from the dependency parser in the Mate tools \cite{Bohnet:2013}, and \emph{contextual} features. The event classification results are modest, reaching only 54.2 F-score for the event class.

The translation model, into which the classifier is integrated, is a 6-gram language model computed over target lemmata using modified Kneser-Ney smoothing and the KenLM toolkit \cite{Heafield:2011}. In addition to the pure target lemma context, it also has access to the identity of the source-language pronoun, used as a concatenated label to each {\sc replace} item. This provides information about the number marking of the pronouns in the source, and also allows for the incorporation of the output of the `it'-label classifier. To predict classes for an unseen test set, a uniform unannotated {\sc replace} tag is used for all classes. The `disambig' tool of the SRILM toolkit \cite{Stolcke:2002} is then used to recover the tag annotated with the correct solution. The combined system with the `it'-labels performed slightly worse than the system without it (57.03 vs.\ 59.84 macro-averaged recall). %The it-label classifier is used only for the English--French task.

The same underlying translation model forms the contrastive system for English--French, and the primary system for all other subtasks.

\subsection{UU-Cap}
The {\sc UU-Cap} approach for English--German uses Conditional Random Fields (CRFs). Pronoun prediction is formulated as a sequence labelling problem, where each word in a sequence is to be labelled as either one of the pronouns or `0' if it does not correspond to a pronoun placeholder. 

This CRF approach has been applied only to German, but there are plans to extend it to other languages.

For German, CRF models are trained using a rich feature set derived from both German and English. The German features include the word sequence itself, the lemma and the PoS-sequence, as well as the gender of the surrounding words (10-gram). The English features include the English word to which the placeholder pronouns have been aligned, and the number and gender features of the surrounding English words (10-gram). 

The CRF model was trained on the IWSLT15 corpus and used the TED talks for development. The rule-based morphological Analyser SMOR \cite{Schmid:2004} as well as its English spinoff EMOR (not published) were used to derive the gender and number of the German and English words.

\subsection{UU-Hardmeier}
The \textsc{UU-Hardmeier} system \cite{UU-Hardmeier:2016} is a system combination of two different models. One of them, based on earlier work \cite{Hardmeier:2013a}, is a feed-forward neural network that takes as input the source pronoun and the source context words, target lemmata and target PoS tags in a window of 3 words to the left and to the right of the pronoun. In addition, the network receives a list of potential antecedent candidates identified by the preprocessing part of a coreference resolution system. Anaphora resolution is treated as a latent variable by the model. This system is combined by linear interpolation with a specially trained 6-gram language model identical to the contrastive system of the \textsc{uuppsala} submission described above. The neural network component on its own was submitted as a contrastive system.

In the evaluation, the system combination of the two components achieved better scores than each component individually. This demonstrates that both components contribute complementary information that is valuable for the task. A rather disappointing result is that the neural network classifier completely fails to predict the rare pronoun classes in this evaluation, even though previous work suggested that this should be one of its strengths \cite{Hardmeier:2013a}. The reasons for this require further investigation.

\subsection{UU-Stymne}
The {\sc UU-Stymne} systems \cite{UU-Stymne:2016} use linear SVM classifiers for all language pairs. A number of different features were explored, but anaphora is not explicitly modelled. The features used can be grouped in the following way: source pronouns, local context words/lemmata, preceding nouns, target PoS \ngrams\ with two different PoS tag-sets, dependency heads of pronouns, target LM scores, alignments, and pronoun position. A joint tagger and dependency parser on the source text is used for some of the features. The primary system is a 2-step classifier where a binary classifier is first used to distinguish between the {\sc other} class and pronoun, then a multi-class classifier distinguishes between the pronoun classes. The secondary system is a standard 1-step classifier. The Mate Tools parser \cite{Bohnet:2012} is used for joint PoS tagging and parsing for all languages.

Across language pairs, source pronouns, local context and dependency features performed best. The LM and preceding noun features hurt performance. For the binary distinction between {\sc other} and pronouns, target PoS \ngrams\ performed well.

The submitted systems for German--English and French--English unfortunately contained a bug in the feature extraction that severely affected the scores. The system description paper also reports the much higher scores with the bug resolved.

\section{Evaluation}
% TO BE COMPLETED BY: Preslav
% LKG: Define the metrics used
% LKG: Justify the use of macro-averaged R, and why this is used instead of macro-averaged F from DiscoMT 2015

% APB: edited the paragraphs below
While in 2015 we used macro-averaged $F_1$ as an official evaluation measure,
this year we adopted \emph{macro-averaged recall}, which was also recently adopted by some other competitions, e.g., by SemEval-2016 Task 4 \cite{SemEval2016-task4}. Moreover, as in 2015, we also report \emph{accuracy} as a secondary evaluation measure.

Macro-averaged recall ranges in $[0,1]$, where a value of 1 is achieved by the perfect classifier,\footnote{If the test data did not have any instances of some of the classes, we excluded these classes from the macro-averaging, i.e., we only macro-averaged over classes that are present in the gold standard.} and a value of 0 is achieved by the classifier that misclassifies all examples. The value of $1/C$, where $C$ is the number of classes, is achieved by a trivial classifier that assigns the same class to all examples (regardless of which class is chosen), and is also the expected value of a random classifier.

The advantage of macro-averaged recall over accuracy is that it is more robust to class imbalance. For instance, the accuracy of the majority-class classifier may be much higher than $1/C$ if the test dataset is imbalanced. Thus, one cannot interpret the absolute value of accuracy (e.g., is 0.7 a good or a bad value?) without comparing it to a baseline that must be computed for each specific test dataset. In contrast, for macro-averaged recall, it is clear that a value of, e.g., 0.7, is well above the majority-class and the random baselines, which are both always $1/C$ (e.g., 0.5 with two classes, 0.33 with three classes, etc.). Standard $F_{1}$ and macro-averaged $F_1$ are also sensitive to class imbalance for the same reason; see \newcite{Sebastiani:2015zl} for more detail.

\section{Results}
% TO BE COMPLETED BY: Preslav
% LKG: Add a very brief introduction to the results, referring to the tables
The results of the evaluation are shown in Tables~\ref{table:results:de-en}-\ref{table:results:en-fr}, one for each subtask. The tables contain two scores: \emph{macro-averaged recall} (the official shared task metric) and \emph{accuracy}.
%(an unofficial, supplementary metric).

As described in Section~\ref{sec:BaselineSystems}, we provide two official baseline scores for each subtask. The first, computed with the {\sc none} penalty set to a default value of zero, appears in the tables as \emph{baseline0}. The second, computed with the {\sc none} penalty set to an optimised value, appears in the tables in the format \emph{baseline$<$penalty$>$}. The optimised penalty values are different for each subtask.

As we use macro-averaged recall as an official evaluation measure,
its value for the majority class and for a random baseline are both $1/C$,
and thus we do not show them in the tables. Specifically, the macro-average recall of the random baseline is 12.50 for English--French and French--English (8 classes each), 20.00 for English--German, and 11.11 for German--English.

\begin{table*}[tbh]
\begin{center}
\begin{tabular}{clcc}
& \bf Submission & \bf Macro-Avg Recall & \bf \scriptsize Accuracy\\
\hline
\bf 1 & \bf TurkuNLP-primary & \bf 73.91$_{1}$ & \bf \scriptsize 75.36$_{3}$ \\
\bf 2 & \bf UKYOTO-primary & \bf 73.17$_{2}$ & \bf \scriptsize 80.33$_{1}$ \\
& TurkuNLP-contrastive & 72.60 & \scriptsize 80.54 \\
\bf 3 & \bf UHELSINKI-primary & \bf 69.76$_{3}$ & \bf \scriptsize 77.85$_{2}$ \\
& UU-Stymne-contrastive & 60.83 & \scriptsize 70.60 \\
\bf 4 & \bf CUNI-primary & \bf 60.42$_{4}$ & \bf \scriptsize 64.18$_{6}$ \\
\bf 5 & \bf UUPPSALA-primary & \bf 59.56$_{5}$ & \bf \scriptsize 73.71$_{4}$ \\
\bf 6 & \bf UU-Stymne-primary & \bf 59.28$_{6}$ & \bf \scriptsize 69.98$_{5}$ \\
& CUNI-contrastive & 56.83 & \scriptsize 65.22 \\
& \it baseline$-1.5$ & \it 44.52 & \it \scriptsize 54.87 \\
& \it baseline0 & \it 42.15 & \it \scriptsize 53.42 \\
\hline
\end{tabular}
\caption{\textbf{Results for German-English.} The first column shows the rank of the primary systems with respect to the official metric: macro-averaged recall. The second column contains the team's name and its submission type: primary vs.\ contrastive. The following columns show the results for each system, measured in terms of macro-averaged recall (official metric) and accuracy (unofficial, supplementary metric). The subindices show the rank of the primary systems with respect to the evaluation measure in the respective column.  The random/majority baseline macro-averaged recall is 11.11.} % APB: added last sentence}
\label{table:results:de-en}
\end{center}
\end{table*}

\textbf{German--English.} 
Table~\ref{table:results:de-en} shows the results for German--English.
We can see that all six participating teams outperform the baselines by a wide margin.
The top systems, {\sc TurkuNLP}, {\sc ukyoto} and {\sc uhelsinki} score between 73.91 and 69.76 in macro-averaged recall.
This is very much above the performance of \emph{baseline0} and \emph{baseline-1.5}, which are in the low-mid 40s. It is also well above the majority/random baseline (not shown) at 11.11, which is outperformed by far by all systems.
Note that the top-3 systems in terms of macro-averaged recall are also the top-3 in terms of accuracy, but in different order.

\begin{table*}[tbh]
\begin{center}
\begin{tabular}{clcc}
& \bf Submission & \bf Macro-Avg Recall & \bf \scriptsize Accuracy\\
\hline
\bf 1 & \bf TurkuNLP-primary & \bf 64.41$_{1}$ & \bf \scriptsize 71.54$_{2}$ \\
& TurkuNLP-contrastive & 58.39 & \scriptsize 72.85 \\
\bf 2 & \bf UKYOTO-primary & \bf 52.50$_{2}$ & \bf \scriptsize 71.28$_{3}$ \\
\bf 3 & \bf UU-Stymne-primary & \bf 52.12$_{3}$ & \bf \scriptsize 70.76$_{4}$ \\
\bf 4 & \bf UU-Hardmeier-primary & \bf 50.36$_{4}$ & \bf \scriptsize 74.67$_{1}$ \\
& UU-Stymne-contrastive & 48.92 & \scriptsize 68.93 \\
\bf 5 & \bf uedin-primary & \bf 48.72$_{5}$ & \bf \scriptsize 66.32$_{6}$ \\
& \it baseline$-2$ & \it 47.86 & \it \scriptsize 54.31 \\
& uedin-contrastive & 47.75 & \scriptsize 64.75 \\
\bf 6 & \bf UUPPSALA-primary & \bf 47.43$_{6}$ & \bf \scriptsize 68.67$_{5}$ \\
& UU-Hardmeier-contrastive & 46.64 & \scriptsize 72.06 \\
\bf 7 & \bf UHELSINKI-primary & \bf 44.69$_{7}$ & \bf \scriptsize 65.80$_{7}$ \\
\bf 8 & \bf UU-Cap-primary & \bf 41.61$_{8}$ & \bf \scriptsize 63.71$_{8}$ \\
& \it baseline0 & \it 38.53 & \it \scriptsize 50.13 \\
& CUNI-contrastive & 30.70 & \scriptsize 46.48 \\
\bf 9 & \bf CUNI-primary & \bf 28.26$_{9}$ & \bf \scriptsize 42.04$_{9}$ \\
\hline
\end{tabular}
\caption{\textbf{Results for English-German.} The first column shows the rank of the primary systems with respect to the official metric: macro-averaged recall. The second column contains the team's name and its submission type: primary vs.\ contrastive. The following columns show the results for each system, measured in terms of macro-averaged recall (official metric) and accuracy (unofficial, supplementary metric). The subindices show the rank of the primary systems with respect to the evaluation measure in the respective column.  The random/majority baseline macro-averaged recall is 20.00.} % APB: added last sentence}
\label{table:results:en-de}
\end{center}
\end{table*}

\textbf{English--German.} 
The results for English--German are shown in Table~\ref{table:results:en-de}.
This direction was arguably harder as about half of the nine participating teams are below the optimised \emph{baseline-2} (with a score of 47.86), and one system is even below \emph{baseline0}. The clear winner is {\sc TurkuNLP}, with a macro-averaged recall of 64.41 (they are also second in accuracy), ahead of {\sc ukyoto} with 52.50 and {\sc UU-Stymne} with 52.12 (third and fourth in accuracy, respectively). All of the systems outperform the majority/random baseline (at 20.00), though some by a smaller margin than for German--English.

\begin{table*}[tbh]
\begin{center}
\begin{tabular}{clcc}
& \bf Submission & \bf Macro-Avg Recall & \bf \scriptsize Accuracy\\
\hline
\bf 1 & \bf TurkuNLP-primary & \bf 72.03$_{1}$ & \bf \scriptsize 80.79$_{2}$ \\
& TurkuNLP-contrastive & 66.54 & \scriptsize 85.06 \\
\bf 2 & \bf UKYOTO-primary & \bf 65.63$_{2}$ & \bf \scriptsize 82.93$_{1}$ \\
\bf 3 & \bf UHELSINKI-primary & \bf 62.98$_{3}$ & \bf \scriptsize 78.96$_{3}$ \\
\bf 4 & \bf UUPSALA-primary & \bf 62.65$_{4}$ & \bf \scriptsize 74.39$_{4}$ \\
& \it baseline$-1.5$ & \it 42.96 & \it \scriptsize 53.66 \\
& \it baseline0 & \it 38.38 & \it \scriptsize 52.44 \\
\bf 5 & \bf UU-Stymne-primary & \bf 36.44$_{5}$ & \bf \scriptsize 53.66$_{5}$ \\
& UU-Stymne-contrastive & 34.12 & \scriptsize 52.13 \\
\hline
\end{tabular}
\caption{\textbf{Results for French-English.} The first column shows the rank of the primary systems with respect to the official metric: macro-averaged recall. The second column contains the team's name and its submission type: primary vs.\ contrastive. The following columns show the results for each system, measured in terms of macro-averaged recall (official metric) and accuracy (unofficial, supplementary metric). The subindices show the rank of the primary systems with respect to the evaluation measure in the respective column.  The random/majority baseline macro-averaged recall is 12.50.} % APB: added last sentence}
\label{table:results:fr-en}
\end{center}
\end{table*}

\textbf{French--English.} 
The results for French--English are shown in Table~\ref{table:results:fr-en}.
Four of the five participating teams had a macro-averaged recall score above 50.00, and outperformed the LM-based baselines at 38.38 and 42.96 for the tuned and the untuned version, respectively.
All of the systems outperformed by far the majority/random baselines at 12.50.
Once again, {\sc TurkuNLP} is the clear winner with 72.03 (second in accuracy).
It is followed by {\sc ukyoto} with 65.63 (first in accuracy),
{\sc uhelsinki} with 62.98 (third in accuracy),
and {\sc uuppsala} with 62.65 (fourth in accuracy).

\begin{table*}[tbh]
\begin{center}
\begin{tabular}{clcc}
& \bf Submission & \bf Macro-Avg Recall & \bf \scriptsize Accuracy\\
\hline
\bf 1 & \bf TurkuNLP-primary & \bf 65.70$_{1}$ & \bf \scriptsize 70.51$_{5}$ \\
\bf 2 & \bf UU-Stymne-primary & \bf 65.35$_{2}$ & \bf \scriptsize 73.99$_{2}$ \\
\bf 3 & \bf UKYOTO-primary & \bf 62.44$_{3}$ & \bf \scriptsize 70.51$_{4}$ \\
\bf 4 & \bf uedin-primary & \bf 61.62$_{4}$ & \bf \scriptsize 71.31$_{3}$ \\
& TurkuNLP-contrastive & 61.46 & \scriptsize 72.39 \\
& UU-Stymne-contrastive & 60.69 & \scriptsize 71.05 \\
\bf 5 & \bf UU-Hardmeier-primary & \bf 60.63$_{5}$ & \bf \scriptsize 74.53$_{1}$ \\
& UUPPSALA-contrastive & 59.84 & \scriptsize 70.78 \\
& uedin-contrastive & 59.83 & \scriptsize 68.63 \\
& limsi-contrastive & 59.34 & \scriptsize 68.36 \\
\bf 6 & \bf limsi-primary & \bf 59.32$_{6}$ & \bf \scriptsize 68.36$_{7}$ \\
\bf 7 & \bf UHELSINKI-primary & \bf 57.50$_{7}$ & \bf \scriptsize 68.90$_{6}$ \\
& \it baseline$-1$ & \it 50.85 & \it \scriptsize 53.35 \\
& UU-Hardmeier-contrastive & 50.80 & \scriptsize 71.31 \\
\bf 8 & \bf UUPPSALA-primary & \bf 48.92$_{8}$ & \bf \scriptsize 62.20$_{8}$ \\
& \it baseline0 & \it 46.98 & \it \scriptsize 52.01 \\
\bf 9 & \bf Idiap-primary & \bf 36.36$_{9}$ & \bf \scriptsize 51.21$_{9}$ \\
& Idiap-contrastive & 30.44 & \scriptsize 42.09 \\
\hline
\end{tabular}
\caption{\textbf{Results for English-French.} The first column shows the rank of the primary systems with respect to the official metric: macro-averaged recall. The second column contains the team's name and its submission type: primary vs.\ contrastive. The following columns show the results for each system, measured in terms of macro-averaged recall (official metric) and accuracy (unofficial, supplementary metric). The subindices show the rank of the primary systems with respect to the evaluation measure in the respective column.  The random/majority baseline macro-averaged recall is 12.50.} % APB: added last sentence
\label{table:results:en-fr}
\end{center}
\end{table*}

\textbf{English--French.} 
The results for English--French are shown in Table~\ref{table:results:en-fr}. Seven of the nine participating teams outperformed the two baselines (in fact, \emph{baseline0} was outperformed by all but one team).
All of the participants outperformed the majority/random baseline of 12.50.
%, as well as the other two baselines.
% Note that this time the tuned LM baseline is slightly better than the majority/random baselines: 50.85\% vs.\ 50.00. % APB: much better than the actual majority F1 of 12.5\%, but is it worth saying it?
The top system is {\sc TurkuNLP} once again, with macro-averaged recall of 65.70, which is barely better than the 65.35 score of {\sc UU-Stymne} (second in accuracy).
The third-best result, 62.44, is that of {\sc ukyoto} (fourth in accuracy).

Overall, there is a clear winner, {\sc TurkuNLP}, which won all four pairs/directions, in two of the cases by a large margin. Naturally, \emph{baseline0} performs worse than the tuned LM baseline in all four cases. 
% More interestingly, both LM baselines are worse than the majority/random baseline (with the exception of one case where the tuned baseline barely passes 50.00). % APB: not true if the baseline is actually 12.5\%
Accuracy scores do not align perfectly well with macro-averaged recall, but the top systems in macro-averaged recall are generally also among the top in terms of accuracy.

% SS: should we have more specific results, especially results for each pronoun? This would mean many big tables, but I think it might be worth it. It could be as an appendix?

\section{Discussion}
% TO BE COMPLETED BY: ???
% LKG: Discuss the results
% LKG: More comparison and contrasting of systems
% LKG: Comments on specific language pairs
% LKG: Generalisations across different language pairs
This year, almost all participating teams managed to outperform the corresponding baselines in their respective subtasks. This applies not only to the majority/random baselines, which proved quite easy to beat, but also to the more sophisticated LM-based baseline with tuned parameters.
This is in stark contrast with the DiscoMT~2015 task, where none of the participating systems was able to outperform the baseline.

In the following subsections, we discuss the success of the WMT~2016 task with respect to the challenges of the individual subtasks, and the design of the submitted systems. We also include a brief comparison with the DiscoMT~2015 task.

\subsection{Challenges}

The subtasks each with different combinations of source-language pronouns and target-language prediction classes, provide different challenges. Judging by the results, the prediction of pronouns for English--French and English--German was more difficult than for the reverse directions. This is perhaps to be expected given the agreement problems associated with predicting the translation of ambiguous English third-person singular pronouns in languages with grammatical gender. However, that is not to say that this is the only problem that these translation directions present.

In the case of English--French translation, systems must accurately determine when to use gendered vs.\ non-gendered translations of anaphoric pronouns. This is in addition to the problems arising from functional ambiguity in the source language. Nevertheless, the English--French and English--German tasks received a greater number of submissions than the tasks for the reverse directions. This is perhaps due to the greater availability of tools and resources for English, than for French and German, coupled with a tendency to focus more on source-language processing.

\subsection{Comparison with the DiscoMT~2015 Task}

The DiscoMT and WMT tasks are not directly comparable. The WMT~2016 baseline, also an \ngram\ language model, is trained on lemmatised, PoS-tagged data, and therefore cannot predict plural pronoun forms. We might therefore consider the WMT~2016 baseline systems to be weaker than the DiscoMT~2015 baseline, which is trained on fully inflected data. However, the submitted systems also have to contend with the same problem of missing number information on target-language nouns and pronouns. The fact that the systems were able to beat the baseline validates the use of more complex features and methods than simply relying on local target-side context.
%and reflects the efforts of the shared task participants.

\subsection{Submitted Systems}

The submitted systems used recurrent neural networks ({\sc TurkuNLP} and {\sc ukyoto}), linear models ({\sc cuni}), including SVMs ({\sc UU-Stymne} and {\sc uhelsinki}), Maximum Entropy classifiers ({\sc uedin}), Conditional Random Fields ({\sc UU-Cap}), random forests ({\sc limsi}), pronoun-aware language models ({\sc idiap} and {\sc uuppsala}), and a system combination incorporating a classifier and language model ({\sc UU-Hardmeier}).

Overall, the most successful systems used recurrent neural networks ({\sc TurkuNLP} and {\sc ukyoto}).
The {\sc TurkuNLP} system, which was the best performing system for all four subtasks, is a deep recurrent neural network, optimised to place a greater emphasis on the rare pronoun classes instead of the most common ones. The authors claim that the English--French and English--German systems in particular benefit from this greater emphasis on rare pronoun classes. However, this is not the only reason for its high performance, as the contrastive system, which treats all pronoun classes equally, also performs well.
%suggesting that the optimisation of the primary system is not the only factor in its success. 
The {\sc ukyoto} team, whose system ranked second in three of the subtasks, report that the system performs well for common pronoun classes but poorly on rare ones, suggesting room for future improvement.

Given the good performance of the two recurrent neural network systems, we might conclude that this architecture is a suitable choice for the cross-lingual pronoun prediction task. It is difficult to determine any further clear patterns in terms of architecture type and performance.

The systems used a wide variety of features, and can be split into two main groups: those that use only contextual information from the source and the target language ({\sc TurkuNLP}, {\sc ukyoto}, {\sc uhelsinki}, and the {\sc uuppsala} source-aware language models), and those that make additional use of external tools and resources ({\sc cuni}, {\sc idiap}, {\sc limsi}, {\sc uedin}, the {\sc uuppsala} primary system for English--French, {\sc UU-Cap}, {\sc UU-Hardmeier} and {\sc UU-Stymne}). 

Popular external tools include those for anaphora/coreference resolution ({\sc cuni}, {\sc limsi} and {\sc uedin}), pleonastic ``it'' detection ({\sc cuni}, {\sc uedin} and {\sc uuppsala}) and dependency parsing ({\sc cuni}, {\sc limsi}, {\sc uuppsala} and {\sc UU-Stymne}). Beyond the observation that recurrent NNs perform well, there seems to be no clear pattern as to whether using external tools and resources vs.\ context only works best. However, context-only methods are applicable to any language pair.

In terms of data, most systems were trained only on the datasets provided for the shared task. The {\sc cuni} system used a wordlist with gender distributions collected by \newcite{bergsma-gennum}, the {\sc idiap} system used the \witiii\ corpus \cite{Cettolo:2012}, and the `it'-disambiguation classifier used in the {\sc uuppsala} system was trained on annotated data from ParCor \cite{Guillou:2014} and the \emph{DiscoMT2015} test set \cite{DiscoMT2015TestSet}.

\section{Conclusions}
% TO BE COMPLETED BY: ???
% LKG: Summarise the main findings
% LKG: Report on the success / failure of the task (pretty sure it was a success!)
We have described the design and the evaluation of the shared task on cross-lingual pronoun prediction at WMT~2016. The task is similar to the DiscoMT~2015 task, which focused on English--French translation. This year, we invited participants to submit systems for four subtasks: for the English--French and English--German language pairs, in both translation directions. Unlike the DiscoMT~2015 task, in which fully inflected target-language sentences were provided in the training and test data, we provided a lemmatised, PoS-tagged representation.

%We prepared and released training and testing datasets, and baseline systems for four subtasks.

We built on the success of the DiscoMT~2015 shared task, attracting increased attention from the community in terms of the number of participants. We received submissions from eleven groups, with many teams submitting systems for several subtasks. This year, the majority of the systems outperformed the official shared task baselines. This is in stark contrast to last year, where none of the systems was able to beat the baseline, an \ngram\ language model. Several factors may have affected this including changes to the task itself, and improved methods.
We hope that the success in the cross-lingual pronoun prediction task will soon translate into improvements in pronoun translation by complete MT pipelines.

\section{Acknowledgements}
The organisation of this task has received support from the following project: Discourse-Oriented Statistical Machine Translation funded by the Swedish Research Council (2012-916).
The work of Chistian Hardmeier and Sara Stymne is part of the Swedish strategic research programme eSSENCE.

\bibliography{acl2016}
\bibliographystyle{acl2016}

%\appendix

\end{document}